\documentclass[sigconf]{acmart}

\usepackage{algorithmic}
\usepackage{algorithm}
\usepackage{amsthm}
\usepackage{amsfonts,dsfont}

\theoremstyle{definition}
\newtheorem{definition}{Definition}[section]
\newcommand{\ind}{\mathds{1}}   %PG use \ind for indicator function
\usepackage{soul}

\DeclareMathOperator{\EX}{\mathbb{E}}% expected value
\DeclareMathOperator{\SIG}{\sum_{k=1}^{\alpha}
    k  B(k)}% sigma
\DeclareMathOperator{\SIGDOLLAR}{(\bar{R}^i)^2\sum_{k=1}^{\alpha}
    \left(B(k)\right)^2}% sigma coin
    
\DeclareMathOperator{\INDEXCOINCIDENCE}{\sum_{k=1}^{\alpha}
    \left(B(k)\right)^2}% sigma coin
    
\DeclareMathOperator{\lvalue}{\frac{2\phi \bar{R}^i \SIG}{\Delta_i} + \frac{4 \ln t\SIGDOLLAR }{\Delta^2_i} \left( 1+ \sqrt{1+ \frac{\Delta_i \phi \SIG}{ \bar{R}^i\ln t \INDEXCOINCIDENCE }}\right)}% l
\DeclareMathOperator{\lvalueleft}{\frac{2\phi \bar{R}^i \SIG}{\Delta_i}}% l
\DeclareMathOperator{\lvalueright}{+ \frac{4 \ln t\SIGDOLLAR }{\Delta^2_i} \left( 1+ \sqrt{1+ \frac{\Delta_i \phi \bar{R}^i \SIG}{\ln t \SIGDOLLAR }}\right)}% l

\DeclareMathOperator*{\argmax}{argmax}
\AtBeginDocument{%
  \providecommand\BibTeX{{%
    \normalfont B\kern-0.5em{\scshape i\kern-0.25em b}\kern-0.8em\TeX}}}

\setcopyright{none}

\newcommand\Citetext[1]{%
  \citeauthor{#1}~[\citeyear{#1}]}

\begin{document}

\title{Generalizing distribution of partial rewards for multi-armed bandits with temporally-partitioned rewards}


\author{Ronald C. van den Broek}
\affiliation{%
  \institution{Eindhoven University of Technology}
  \city{Eindhoven}
  \country{The Netherlands}}
\email{r.c.v.d.broek@student.tue.nl}

\author{Rik Litjens}
\affiliation{%
  \institution{Eindhoven University of Technology}
  \city{Eindhoven}
  \country{The Netherlands}}
  \email{r.litjens@student.tue.nl}
 
\author{Tobias Sagis}
\affiliation{%
  \institution{Eindhoven University of Technology}
  \city{Eindhoven}
  \country{The Netherlands}}
\email{t.g.m.sagis@student.tue.nl}

\author{Luc Siecker}
\affiliation{%
  \institution{Eindhoven University of Technology}
  \city{Eindhoven}
  \country{The Netherlands}}
\email{l.r.siecker@student.tue.nl}

\author{Nina Verbeeke }
\affiliation{ 
  \institution{Eindhoven University of Technology}
  \city{Eindhoven}
  \country{The Netherlands}}
\email{ n.c.verbeeke@student.tue.nl }

 \author{Pratik Gajane}
\affiliation{ 
  \institution{Eindhoven University of Technology}
  \city{Eindhoven}
  \country{The Netherlands}}
\email{ p.gajane@tue.nl }
% \renewcommand{\shortauthors}{Group 5 member et al.}

\acmConference{29th ACM SIGKDD}{August 2023}{t.b.a.}

\begin{abstract}
We investigate the Multi-Armed Bandit problem with Temporally-Partitioned Rewards (TP-MAB) setting in this paper. In the TP-MAB setting, an agent will receive subsets of the reward over multiple rounds rather than the entire reward for the arm all at once. 
% In this paper we have an investigation into how an existing formulation of distributing rewards can be generalized.
%This research builds on previous work that introduced the $\alpha$-smoothness property for characterizing the reward structure. 
%In this paper, this property will be replaced by a more general formulation of how an arm's cumulative reward is distributed across several rounds.
In this paper, we introduce a general formulation of how an arm's cumulative reward is distributed across several rounds, called $\beta$-spread property.
Such a generalization is needed to be able to handle partitioned rewards in which the maximum reward per round is not distributed uniformly across rounds. We derive a lower bound on the TP-MAB problem under the assumption that $\beta$-spread holds. Moreover, we provide an algorithm \texttt{TP-UCB-FR-G}, which uses the $\beta$-spread property to improve the regret upper bound in some scenarios. By generalizing how the cumulative reward is distributed, this setting is applicable in a broader range of applications.
\end{abstract}


% Disable ACM reference format
\settopmatter{printacmref=false,printfolios=true}

% Removes footnote with conference information in first column
\renewcommand\footnotetextcopyrightpermission[1]{}

\maketitle

\section{Introduction}
The multi-armed bandit (MAB) is a framework within reinforcement learning to model sequential decision-making. In the classic MAB problem, an agent is faced with a finite set of $k$ different actions, known as arms. At each point in time, the agent selects one arm and, after pulling it, observes a reward from that arm. The reward is drawn from an arm-specific probability distribution, which is initially unknown to the agent. Given a finite number of $n$ rounds, the agent faces a trade-off between the exploitation of the arm with the highest expected reward and exploration to learn more about the expected rewards of the other arms. The objective of the agent is to accumulate as much reward as possible over $n$ rounds. The sum of the rewards collected over $n$ rounds is commonly referred to as a cumulative reward. To maximize the cumulative reward, the agent must minimize total regret. The total regret is the expected regret of not pulling the best arm.\\
Multi-arm bandit literature typically focuses on scenarios where the rewards are assumed to arrive immediately after performing an action. However, in many real-world scenarios, there is a delay between the execution of an action and the observation of its reward. This has been studied in classical delayed-feedback bandits [\citenum{Joulani2013}, \citenum{mandel2015queue}, \citenum{pike2018bandits}]. In those studies, the reward is assumed to be concentrated in a single round that is delayed. This setting can be expanded by allowing the reward to be partitioned into partial rewards that are observed with different delays. This type of bandit problem, known as MAB with Temporally-Partitioned Rewards (TP-MAB), was introduced in a paper by \Citetext{Romano2022}.  \\
In the TP-MAB setting, the reward from an arm is distributed over multiple rounds. This means that instead of receiving the entire reward for the arm at once, an agent will receive subsets of the reward over multiple rounds. The per-round reward is defined as the partial reward observed by the agent in a single round. It is assumed that the per-round reward is the realization of a random variable with an unknown probability distribution. The cumulative reward of an arm is the sum of all the per-round rewards obtained by pulling an arm. The maximum delay until receiving the cumulative reward of an arm is assumed to be finite. \\
\Citetext{Romano2022} present $\alpha$-smoothness, a property that defines how an arm's cumulative reward is distributed across several rounds. This property examines the reward seen across a group of $\phi$ rounds. If the $\alpha$-smoothness holds, the total reward seen in this group cannot be greater than a fraction of an arm's maximum cumulative reward. The fraction is determined by the value of $\alpha$. The total cumulative reward divided by $\alpha$ is the maximum total reward observed in a group of $\phi$ rounds. \\
Consider the following example to demonstrate $\alpha$-smoothness further. An agent needs to sell a certain product and is given a range of possible prices for that specific product. The agent should select a price, which corresponds to pulling an arm, to sell the product in the following month in a specific affiliate. The selected time horizon to sell the items is set to 30 days, and one round is equal to 1 day. Assume there are 300 items with a fixed price of three euros. The total cumulative reward is 900 euros. If no assumptions are made about how the whole reward will be distributed across the rounds, it is possible that all products may be sold on the final day. As a result, the setting is identical to the classic delayed feedback setting. According to \Citetext{Romano2022}, it is not possible to construct algorithms with regret upper bounds that are better than those of the algorithms for the delayed-feedback situation. \\
To be able to construct better regret bounds, \Citetext{Romano2022} make an assumption that $\alpha$-smoothness holds. In the example described above, the group size $\phi$ might be set to five days, the $\alpha$-smoothness property will specify the maximum reward that may be observed in those five days. If we set $\alpha$ to 6, then the total reward which can maximum received in 5 days is 150 euros. By assuming $\alpha$-smoothness with $\alpha > 1$, the scenario in which an arm's whole reward is obtained in the last round is not possible anymore. By characterizing the reward structure, \Citetext{Romano2022} succeed in providing algorithms that provide better asymptotic regret upper bounds than delayed-feedback bandit algorithms. \\
The $\alpha$-smoothness property states that the maximum reward in a group of consecutive rewards cannot exceed a fraction of the maximum reward. However, in many scenarios, there might be additional information available about how the cumulative reward is spread over the rounds. For example, the majority of the cumulative reward is observed in the first rounds, and the observed partial reward declines exponentially after that. The assumption of $\alpha$-smoothness does not fit well if the cumulative reward is not uniformly spread. This paper will introduce a more generalized way of formulating how an arm's cumulative reward is distributed across several rounds. The scenarios where the cumulative reward is not uniformly spread over the rounds will be investigated. This will be studied by allowing the cumulative reward spread to be any function, such as a curve. \\
A motivating application could be on websites that provide Massive Open Online Courses (MOOCs). The most well-known MOOC providers are Coursera, Udacity, the Khan Academy, and edX. MOOC providers want to be able to provide students with useful recommendations for courses. This problem can be modeled as a TP-MAP problem. A course, which consists of a series of video lectures, might be thought of as an arm. A course can be recommended to a student by an agent, which corresponds to pulling an arm. When the student follows a course, the agent can observe partial rewards (e.g. by asking for a course rate or by checking the watch time retention). The cumulative reward is the sum of rewards from all video lectures in a course. If the $\alpha$-smoothness holds, the total reward observed in a collection of $\phi$ video lectures cannot be more than a fraction of the maximum cumulative reward of the course. However, many students watch the video lectures at the beginning of a course but never get to the course's last few video lectures. As a result, the assumption of the $\alpha$-smoothness property does not really fit in such a case. Therefore, it might be interesting to investigate a more generalized way of formulating how an arm's cumulative reward is distributed across several rounds. \\
The key contribution of this work is that it provides a more generalized way of describing how an arm's cumulative reward is distributed across rounds. We introduce the $\beta$-spread property to replace the $\alpha$-smoothness property. 
%The TP-MAP problem can have a reduced lower bound in some scenarios under this new assumption.
We also propose an algorithm \texttt{TP-UCB-FR-G} that uses this new property. Similarly, tighter upper bounds can be achieved for this algorithm compared to \texttt{TP-UCB-FR} \cite{Romano2022}  under certain assumed distributions.

\begin{figure}[h]
  \includegraphics[width=1\linewidth]{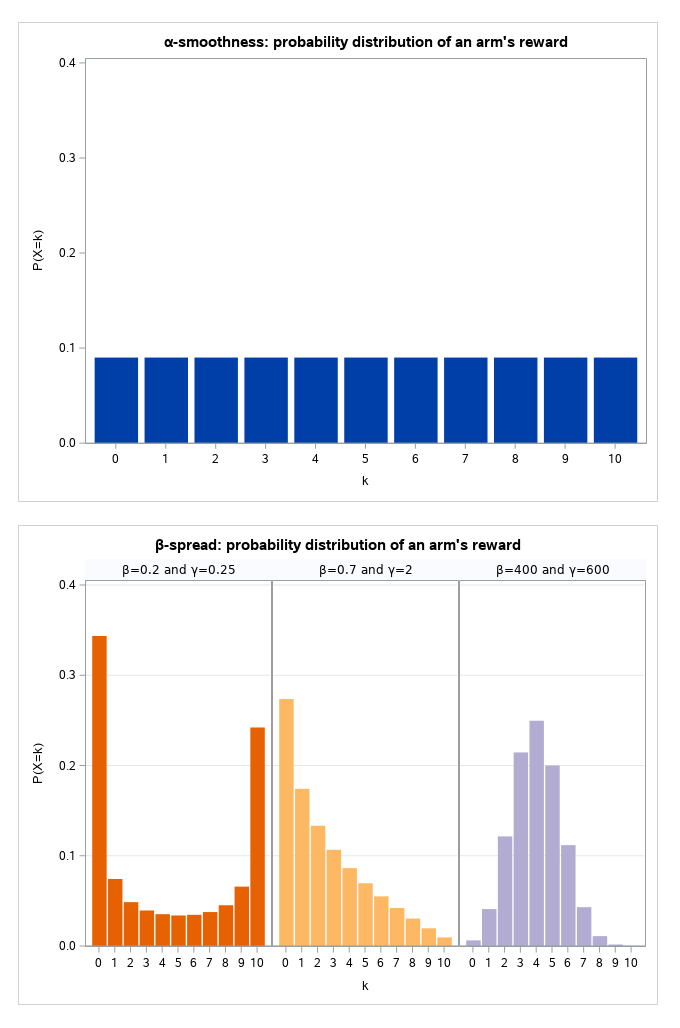}
  \caption{Probability distribution of $\alpha$-smoothness and possible $\beta$-spread probability distributions}
  \label{fig:BBD}
\end{figure}

\section{Background and Related Work}
The classical multi-armed bandits (MAB) problem is the formulation of the exploitation-exploration dilemma inherent to reinforcement learning \cite{DBLP:journals/ftml/BubeckC12}. 
%In this problem, the learner has access to a number of available actions (symbolized by arms) and she has to select one of them (symbolized by pulling an arm), over a period of time, to maximize her cumulative reward.  
In this problem, the learner has to choose between $K$ different actions (also called by arms).
%This problem is formulated as making a choice between $K$ different actions, which can determine a specific reward for the choice you have made. 
A choice is made in every round and is represented by pulling one of the $K$ arms. The successive plays of an arm are independent and identically distributed \cite{Auer2002}. We assume that reward for an arm $i$ is drawn from an unknown distribution on $[0,1]$ with mean $\mu_i$.
%The algorithm that chooses which arm to play next can have an allocation strategy or policy. This policy can determine how an algorithm should choose based on past plays and their associated rewards. This feedback can also be used to determine the regret of such an algorithm. 
Regret is defined as the expected loss, due to the fact that a policy is not making the optimal choice. The regret function of an algorithm \texttt{A} after $T$ time steps is defined as follows \cite{Auer2002}:

$$
\mu^*T - \mu_j \sum_{j=1}^K \mathbb{E}\left[N_j(T)\right] \quad \text { where } \mu^* \stackrel{\text { def }}{=} \max _{1 \leq i \leq K} \mu_i
$$

The goal of algorithm \texttt{A} is to make sure that the received reward will be maximized and therefore minimize the regret function. To make sure this is achieved, it is important for a policy to make sure it explores potentially better arms but also exploits the currently known good choices.
%In order for the agent to get the maximum reward with a specific policy, it is important to make a policy that explores all the possible rewards but also exploits the currently known rewards.
Knowing this goal, the arms can be chosen optimistically via the Upper Confidence Index. 
%Using the index, the upper confidence bound can be determined. 
Based on this idea, \Citetext{Auer2002} proposed an algorithm called Upper Confidence Bound (UCB1) and proved that it has logarithmic regret.
%and that no need for knowledge about the reward distribution is required, except for the fact that the random variable is in the sample space [0,1]. 
%\texttt{UCB1} maximizes the following formula: 
%$x_j + \sqrt{2\ln (n)/n_j}$.
%In this formula, $x_j$ is the average reward by an arm $j$, $n_j$ is the times the arm has played $j$ and $n$ is the total number of plays. \cite{Auer2002}
\\\\
The \texttt{UCB1} algorithm \cite{Auer2002} can be extended to work with delayed rewards. The seminal work of \Citetext{Joulani2013} proposes the concept of a partial monitoring setting, covering all previous, less general, work regarding feedback delays. In the delayed model, feedback on a decision made in timestep $t$ will be received at time step $t+\tau_t$ after the decision is made. The number of delayed time steps is denoted by $\tau_t$. Note that using $\tau_t \equiv 0$ corresponds to the non-delayed model with instant feedback.
%To implement delayed feedback \Citetext{Joulani2013} propose the \texttt{Delayed-UCB1} algorithm, with the same regret guarantees compared to their non-delayed version, with an additive penalty depending on the number of delays in a stochastic setting.
%\\\\
However, in real-world applications, it is common that the rewards that are received from pulling an arm are spread over an interval. These rewards are convoluted with each other \cite{Wang2021}. Underestimating the intervals leads to no theoretical result, and overestimating results in a worse performance \cite{Wang2021}. Past solutions required knowledge about the delay of the reward, which results in a non-anonymous setting \cite{Joulani2013}. This is not realistic, so \Citetext{Wang2021} demonstrated a method to establish an upper regret bound for the algorithm they use, removing the need for prior knowledge.
\\\\
\Citetext{Romano2022} introduce a novel bandit setting, namely Multi-Armed Bandit with Temporally-Partitioned Rewards (TP-MAB). In this setting, a stochastic reward that is received by pulling an arm is partitioned over a finite number of rounds followed by the pull. In this study, two algorithms were designed: \texttt{TP-UCB-FR} and \texttt{TP-UCB-EW}. Both algorithms that were introduced used temporally spaced rewards, and it is shown that the upper regret bound for these two algorithms is respectively $\mathcal{O}(\ln T/\alpha)$ and $\mathcal{O}(\ln T)$, for \texttt{TP-UCB-FR} and \texttt{TP-UCB-EW} \cite{Romano2022}. \texttt{TP-UCB-FR} is an algorithm that uses UCB and Fictitious Realizations. In addition, this algorithm uses an $\alpha$-property that is defined by the researchers. At each point in time ${t \in \{1,...,T\}}$ the agent pulls an arm ${i \in [K]}$. Subsequently, at each round $m$ the agent received a per-round reward $x_{t,m-t+1}^{i}$. It is known to the agent which arm pull produced this reward. The time span over which the reward is partitioned is denoted by $\tau_{max}$. In the definition of $\alpha$-smoothness, the value of $\alpha$ is considered to be a factor of  $\tau_{max}$, thus $\frac{\tau_{max}}{\alpha} := \phi$. The random variable $Z_{t,k}^i$ corresponds to the total reward the agent has observed for pulling arm $i$ at time $t$, over the duration of $\phi$ consecutive rounds in group $k$. The group index $k$ denotes $(k-1)\phi$ rounds of observations that have already passed. It is defined as the sum of per-round rewards. $Z_{t,k}^i := \sum_{j=(k-1)\phi+1}^{k\phi} X_{t,j}^i $. 
The value of $\alpha$ lies between 1 and $\tau_{max}$, thus ${\alpha \in \{1,...,\tau_{max}\}}$. \Citetext{Romano2022} define $\alpha$-smoothness as follows:
\begin{definition} ($\alpha$-smoothness) For $\alpha \in \{1,...,\tau_{max}\}$, the reward is $\alpha$-smooth iff $\frac{\tau_{max}}{\alpha}$ with $\phi \in \mathbb{N}$ and for each $k \in \{1,...,K\}$ the random variables $Z_{t,k}^i$ are independent and s.t. $\overline{Z}_{\alpha,k}^i = \overline{Z}_{\alpha}^i = \frac{\overline{R}^i}{\alpha}$.
\end{definition}
Because of the use of this specific property, there is no general function that can work in a broader variety of settings. As a remedy, we propose to use general distributions that can more accurately characterize how the received reward is partitioned. An example is a Beta-Binomial distribution. This probability distribution uses two parameters to determine its shape, and this can be implemented such that it determines the distribution of a received reward over time. Examples of this distribution can be seen in figure \ref{fig:BBD}. Other discrete distributions over a finite domain can also be used, such as binomial distributions. 
%However, the Beta-Binomial distribution uses a random probability of success at each round instead of a fixed parameter, which probably will suit our application of MAB better. Furthermore, a Beta-Binomial distribution does depend on three parameters, namely the number of trials $n$, $\alpha$, and $\beta$. The latter two parameters determine the shape of the distribution and need to be non-negative integers \cite{Rodriguez-Avi2007}.

\section{Problem Formulation}
Consider a MAB problem with $K$ arms over a time horizon of $T$ rounds, where $K, T \in \mathbb{N}$. At every round $t \in \{1, 2, ..., T\}$ an arm $i \in \{1, 2,...,K\}$ is pulled. Unlike the delayed-reward bandit problem, the total reward is temporally partitioned over the set of rounds $T' = \{t, t+1,...,\tau_{max}\}$. Let $x^i_{t, m} (m \in T')$ denote the partitioned reward that the learner receives at round $m$, after pulling the arm $i$ at round $t$. The cumulative reward is completely collected by the learner at round $\tau_{max}$. Each per-round reward $x^i_{t, m}$ is represented as the realization of a random variable $X^i_{t, m}$ with support $[0,  \overline{X}^i_{m}]$\footnote{Without loss of generality, we assume that the per-round rewards $X^i_m \geq 0$, and trivially, the same condition holds for the cumulative reward over $|T'|$ rounds ($R^i_t$)}. The cumulative reward collected by the learner from pulling arm $i$ at round $t$ is denoted by $r^i_t$ and it is the realisation of a random variable $R^i_t$ such that $R^i_t := \sum_{n=t}^{\tau_{max}} X^i_{t, n}$ with support $[0, \overline{R}^i]$. Straightforwardly, we observe that $\overline{R}^i := \sum_{n=t}^{\tau_{max}} \overline{X}^i_{n}$.\\
\Citetext{Romano2022} have shown that, in practice, per-round rewards for an arm provide information on the cumulative reward of the arm. The authors introduce an $\alpha$-smoothness property that partitions the temporally-spaced rewards such that each partition corresponds to the sum of a set of consecutive per-round rewards. Formally, let $\alpha \in T'$ be such that $\alpha$ is a factor of $(\tau_{max}-t)$. The cardinality of each partition, which we shall name `z-group' from now on, is denoted by $\phi := \frac{(\tau_{max}-t)}{\alpha}$ with $\phi \in \mathbf{N}$. Similar to above, we can now define each z-group $z^i_{t, k}, k \in \{1, 2, ..., \alpha\}$ as the realisation of a random variable $Z^i_{t, k}$ such that for every $k$:
\begin{equation}
    Z^{i}_{t, k} := \sum_{n=t+(k-1)\phi}^{t+k\phi-1} X^{i}_{t, n}
\end{equation}
Finally, we note that $Z^i_{t, k}$ has support $[0, \overline{Z}^i_{\alpha, k}]$. We can distinguish two cases when considering extreme values of $\alpha$:
\begin{enumerate}
    \item \textbf{Case: $\alpha = 1$}: All temporally-spaced rewards are placed in a single z-group over a time horizon.
    \item \textbf{Case: $\alpha = \tau_{max}$}: The number of z-groups equals the number of temporally-spaced rewards over a time horizon.
\end{enumerate}

% \\ With the $\alpha$-smoothness property, we can further tighten the lower regret bounds proposed in the work by \Citetext{Romano2022}. 
The regret of any uniformly efficient policy $\mathfrak{U}$ applied to the assumption-free TP-MAB problem has a lower bound of \cite{Romano2022}:

\begin{equation}
    \lim \inf _{T \rightarrow+\infty} \frac{\mathcal{R}_T(\mathfrak{U})}{\ln T} \geq \sum_{i: \mu_i<\mu^*} \frac{\Delta_i}{K L\left(\frac{\mu_i}{\bar{R}_{\max }}, \frac{\mu^*}{R_{\max }}\right)},
\end{equation}

In \Citetext{Romano2022}, the authors manage to tighten this lower bound by assuming $\alpha$-smoothness. The attained bound is given by:

\begin{equation}
\label{eq:loweralphasmooth}
        \lim \inf _{T \rightarrow+\infty} \frac{\mathcal{R}_T(\mathfrak{U})}{\ln T} \geq \sum_{i: \mu_i<\mu^*} \frac{\Delta_i}{\alpha K L\left(\frac{\mu_i}{\bar{R}_{\max }}, \frac{\mu^*}{R_{\max }}\right)},
\end{equation} \\
The $\alpha$-smoothness property ensures that all temporally-partitioned rewards contribute towards bounding the values of future rewards within the same window. 
% \Citetext{Romano2022} assume an equal probability of attaining a partial reward to every z-group, thereby creating upper bounds of:
% \begin{equation}
%     \overline{Z}_{\alpha, k}^i=\overline{Z}_\alpha^i=\frac{\overline{R}^i}{\alpha}
% \end{equation}
If the $\alpha$-smoothness holds, then the maximum cumulative reward in a z-group $\overline{Z}_{\alpha, k}^i$ is equal for all z-groups $k \in \{1, 2, ..., \alpha\}$. Therefore, we can say that $\forall_k \in \{1, 2, ..., \alpha\}$, $\overline{Z}_{\alpha, k}^i = \overline{Z}_\alpha^i $. \\ However, the assumption of $\alpha$-smoothness does not fit well if the cumulative reward is not evenly distributed across rounds. The goal of this paper is to \textit{generalize} the spread of the rewards across z-groups.
% such that it can be adjusted accordingly while maintaining the same regret bounds provided in the paper of \Citetext{Romano2022}.
To accomplish this, it should be investigated whether $\alpha$-smoothness can be replaced with a property that allows for the modeling of scenarios in which the cumulative reward is not distributed uniformly across rounds. In other words, we want to eliminate the assumption that every z-group has an equal probability of attaining a partial reward. Thus, we want to define a property that allows $\overline{Z}_{\alpha, k}^i$ to differ across z-groups $k \in \{1, 2, ..., \alpha\}$.

\section{Solution approach: $\beta$-spread property}
The goal of this paper is to eliminate the assumption that every z-group has an equal probability of attaining a partial reward. Instead, the probability of partial reward occurring in a specific z-group will be represented by a discrete function over all possible values $k \in \{1, 2, ..., \alpha\}$ that gives probabilities to partial rewards being obtained in z-group $Z_{\alpha, k}^i$ for all $i$. We introduce an arbitrary discrete probability distribution $\mathbb{F}$ to describe the shape of this function. More formally:

\begin{definition}[$\beta$-spread]
Given a set of values $k \in \{1, 2, ..., \alpha\}$ with corresponding z-groups $Z^i_{\alpha, k}$ for every arm $i \in \{1, 2, ..., K\}$, and an arbitrary discrete probability distribution $\mathbb{F}$ with a finite domain of size $\alpha$, $\beta$-spread is defined as the set of weights $\{P(X = 1), P(X = 2), ..., P(X = \alpha)\}$ such that for each $Z^i_{\alpha, k}$, a unique corresponding weight $P(X = k)$ exists. Let us define a random variable $Y \sim \mathbb{F}$.
\end{definition}

The probability of a partial reward being observed in a specific z-group is formalized by any arbitrary probability distribution of choice. For each reward point, the value of $Y \sim \mathbb{F}$ determines the z-group in which it is observed. As a result, every z-group can now contain a different amount of the maximum partial reward. Let $B(x)$ be the \textit{probability mass function} of the chosen distribution $\mathbb{F}$. We define the new upper bounds for the z-groups for every $k, i$ as:
\begin{equation}
    \overline{Z}_{\alpha, k}^i=\mathrm{B}(k)\cdot\overline{R}^i
\end{equation}
Using this result, we can revise the original lower bound attained by \Citetext{Romano2022}, as stated in Theorem 4.1.

\begin{theorem}
The regret of any uniformly efficient policy $\mathfrak{U}$ applied to the TP-MAB problem with the $\beta$-spread property is bounded from below by:
\begin{equation}
\begin{aligned}
    &\lim \inf _{T \rightarrow+\infty} \frac{\mathcal{R}_T(\mathfrak{U})}{\ln T} \\
    &\geq \sum_{i: \mu_i<\mu^*}
    \frac{2}{(\alpha + 1)}\SIG\cdot\alpha\INDEXCOINCIDENCE
    \frac{\Delta_i}
     {\alpha K L\left(\frac{\mu_i}{\bar{R}_{\max }}, \frac{\mu^*}{R_{\max }}\right)}
\end{aligned}
\end{equation}
\label{betalower}
\end{theorem}

Note that this lower bound resolves to the lower bound of (\ref{eq:loweralphasmooth}) in case of $\alpha$-smoothness. The lower bound for the $\beta$-spread setting is lower than the one in the $\alpha$-smoothness setting when $$\frac{2}{(\alpha + 1)}\SIG\cdot\alpha\INDEXCOINCIDENCE < 1$$ The lower bound is higher in the case this value exceeds 1. The proof of this theorem is given in the next section.

\subsection{$\beta$-spread regret lower bound}
\begin{proof}
Theorem 2.2 in the work by \Citetext{Bubeck-Bianchi2012} gives the framework for this proof. We extend the version by \Citetext{Romano2022} by considering that, as the $\beta$-spread property does not modify the cardinality $\phi$ of the z-groups, we can generalize to a setting where multiple rewards are earned by a single arm pull.\\
Let us define an auxiliary TP-MAB setting where:
\begin{itemize}
    \item only two arms exist with expected values $\mu_1$ and $\mu_2$ s.t. $\mu_2$ < $\mu_1$ < 1
    \item upper bound on the reward for each arm is equal to the maximum upper bound, i.e. $\overline{R}^i_t = \overline{R}_{\max}$
    \item The total rewards in each z-group, $Z^i_{t,k}$, are independent and the expected value of the rewards in each z-group is $\mathrm{B}(k) \cdot \mu_i$.
    \item Total reward in z-group, $Z^i_{t,k}$, is a scaled Bernoulli random variable s.t. $Z^i_{t,k}\in\{0, \mathrm{B}(k) \cdot \overline{R}_{\max}\}$
    %For the sake of argument,%
    \item Pulling an arm at time $t$ provides rewards $\{Z^i_{t,1}, ..., Z^i_{t,\alpha}\}$ that can all be observed immediately at the time of the pull.
\end{itemize}
In this proof of the lower bound, we trivially observe that finding the optimal arm in a setting in which all of the partial rewards are observed at once can never be more difficult than in a setting in which rewards are spread out over a set of rounds $\{t, t+1, ..., \tau_{max}\}$. Therefore, a lower bound in this defined setting gives a lower bound in our $\beta$-spread setting.
\\
To give an idea of how good an arm is compared to its maximum, we derive a new alternative mean for each arm as $\mu_{A_i}=\frac{\mu_i}{\overline{R}_{\max}}$. Note that $\mu_{A_i} < 1$ as mentioned before. 
\\
Let $\mathbb{E}[\mathrm{N}_i(T)]$ denote the \textit{expected} number of times an arm $i$ is pulled over a set of rounds $T$. To compute $\mathbb{E}[\mathrm{N}_1(\{t, t+1, ..., \tau_{max}\})]$ and $\mathbb{E}[\mathrm{N}_2(\{t, t+1, ..., \tau_{max}\})]$, we can use the scaled reward values without loss of validity.
\\
If we now consider a second, modified instance of the above TP-MAB setting, with the only difference being that arm 2 is now the optimal arm s.t. $\mu_{A_1} < \mu'_{A_2} < 1$, we can show that the forecaster choosing the arms cannot distinguish between the different instances. This reasoning implies a lower bound on the number of times a sub-optimal arm is played. We know that $x \mapsto KL(\mu_{A_1}, x)$ is a continuous function and we can find a $\mu'_{A_2}$ for each $\epsilon > 0$, such that:

\begin{equation}
    \label{eq:klinequality}
    KL(\mu_{A_2}, \mu'_{A_2}) \leq (1+\epsilon) KL(\mu_{A_2}, \mu_{A_1})
\end{equation}
\\\\
The proof follows the steps given in the work by \Citetext{Bubeck-Bianchi2012} to derive a lower bound for any uniform policy $\mathfrak{U}$.

\subsection*{Step 1: $\mathbb{P}(C_t) = o(1)$}

For this proof, we change the notation of the rewards slightly such that each variable in the sequence $Z^i_{1, 1}, ..., Z^i_{n, \alpha}$ represents the cumulative reward of an arm $i$ when pulled $n$ times, at timestep $k \in \{1, 2, ..., \alpha\}$ and thus $Z^i_{s, k}$ for $s \in \{1, 2,...,n\}$ represents the cumulative reward of an arm $i$ after the $s$'th pull at timestep $k$. Using this notation, we can define the empirical estimate of $\mathrm{KL}(\mu_{Z_2}, \mu'_{Z_2})$ as:
\begin{equation}
    \widehat{\mathrm{KL}_{s\beta}} := \sum_{n=1}^s\sum_{k=1}^\alpha \mathrm{ln} \frac{\mu_{A_2}Z^2_{n, k} + (1-\mu_{A_2})(1-Z^2_{n, k})}{\mu'_{A_2}Z^2_{n, k} + (1-\mu'_{A_2})(1-Z^2_{n, k})}
\end{equation}
\\\\ 
Using this, we define an event that links the behaviour of the original agent to the modified version

\begin{equation}
    C_t:=\left\{\alpha N_2(t)<f_t \quad \text { and } \quad \widehat{K L}_{\alpha N_2(t)} \leq(1-\varepsilon / 2) \ln t\right\}
\end{equation}
\\
with 
\\
\begin{equation}
    f_t = \left(\frac{2}{\alpha + 1}\SIG \cdot\alpha\INDEXCOINCIDENCE  \right)\frac{1-\epsilon}{KL(\mu_{A_2}, \mu'_{A_2})} \ln t
\end{equation}
\\
Using the change of measure identity defined in \Citetext{Bubeck-Bianchi2012} and the second inequality in $C_t$:

\begin{equation}
    \mathbb{P}^{\prime}\left(C_t\right)=\mathbb{E}\left[1_{C_t} \exp \left(-\widehat{K L_{\alpha N}(t)}\right)\right] \geq e^{-(1-\varepsilon / 2) \ln t} \mathbb{P}\left(C_t\right),
\end{equation}

For the next step we use the fact that $\mathbb{P}^{\prime}\left(C_t\right) \leq \mathbb{P}^{\prime}\left(\alpha N_2(t)<f_t\right)$, Markov's inequality and the fact that the policy $\mathfrak{U}$ is uniformly efficient (i.e. $\EX[N_2(t)] = o(t^{\gamma})$ with $\gamma<1$).

\begin{equation}
    \mathbb{P}\left(C_t\right) \leq t^{(1-\varepsilon / 2)} \mathbb{P}^{\prime}\left(C_t\right) \leq t^{(1-\varepsilon / 2)} \mathbb{P}^{\prime}\left(\alpha N_2(t)<f_t\right) 
\end{equation}   
    
\begin{equation}
    \leq t^{(1-\varepsilon / 2)} \frac{\mathbb{E}^{\prime}\left[t-N_2(t)\right]}{t-f_t / \alpha}=o(1)
\end{equation}

\subsection*{Step 2: $\mathbb{P}\left(\alpha N_2(t)\leq f_t\right)=o(1)$}

Again using Theorem 2.2 from \Citetext{Bubeck-Bianchi2012} and observing that we always have:
\begin{enumerate}
    \item $\SIG \geq 1 \implies \frac{2}{\alpha + 1}\SIG \geq \frac{2}{\alpha + 1}$
    \item $\INDEXCOINCIDENCE \in [\frac{1}{\alpha},1]\implies \alpha\INDEXCOINCIDENCE \in [1,\alpha]$
    \item $\frac{2}{\alpha + 1}\SIG\cdot\alpha\INDEXCOINCIDENCE \geq \frac{2}{(\alpha+1)} > 0$
\end{enumerate}

we obtain:
\begin{align*}
    &o(1)=\mathbb{P}\left(C_t\right) \leq \mathbb{P}(\underbrace{\alpha N_2(t)<f_t}_{E_1}) \wedge\\
   &  
   \frac{\alpha + 1}{2\alpha} \cdot \frac{1}{\SIG \cdot \INDEXCOINCIDENCE} \cdot
   \frac{K L\left(\mu_{Z_2}, \mu_{Z_2}^{\prime}\right)}{(1-\varepsilon) \ln t} \cdot \max_{s<f_t / \alpha} \widehat{K L}_{\alpha s} \\
  & \leq \underbrace{\frac{1-\varepsilon / 2}{1-\varepsilon} \cdot 
  \frac{\alpha + 1}{2\alpha} \cdot \frac{1}{\SIG \cdot \INDEXCOINCIDENCE} \cdot
   K L\left(\mu_{Z_2}, \mu_{Z_2}^{\prime}\right)
   }_{E_2} 
\end{align*}

Using the strong law of large numbers for the event $E_2$ s.t. $\lim _{t \rightarrow+\infty} \mathbb{P}\left(E_2\right)=1$, we conclude that $\mathbb{P}\left(E_1\right)=\mathbb{P}\left(\alpha N_2(t)<f_t\right)=o(1)$, and that for $t \rightarrow+\infty$ we have $\mathbb{E}\left[N_2(t)\right]>f_t / \alpha$.

\subsection*{Final step: deriving the lower bound}

Using Equation (\ref{eq:klinequality}) we know that, for $t \rightarrow+\infty$ :
$$
\mathbb{E}\left[N_2(t)\right]>f_t / \alpha=
\frac{2}{\alpha + 1}\SIG\cdot\alpha\INDEXCOINCIDENCE
\frac{1-\varepsilon}{\alpha K L\left(\mu_{A_2}, \mu_{A_2}^{\prime}\right)} \ln t $$

$$\geq\frac{2}{\alpha + 1}\SIG \cdot \alpha\INDEXCOINCIDENCE
\frac{1-\varepsilon}{\alpha(1+\varepsilon) K L\left(\mu_{A_2}, \mu_{A_1}\right)}
\ln t
$$
where the theorem statement follows from the arbitrarity of the value of $\varepsilon$ \cite{Romano2022}, substituting $\mu_{A_1}$ with $\frac{\mu^*}{\bar{R}_{\max }}$ and $\mu_{A_2}$ with $\frac{\mu_2}{\bar{R}_{\max }}$, and summing over all the suboptimal arms.

\end{proof}
% ...

\section{Algorithm for the $\beta$-spread TP-MAB setting}
In this section, we propose an algorithm that makes use of the $\beta$-spread property in the TP-MAB setting. \texttt{TP-UCB-FR-G} is an extension of the algorithm \texttt{TP-UCB-FR} found in \Citetext{Romano2022}.

\subsection{TP-UCB-FR-G}

\begin{algorithm}
\caption{TP-UCB-FR-G}\label{alg:fr}
\begin{algorithmic}[1]
    \STATE $\textbf{Input: } a \in [\tau_{max}], \tau_{max} \in \mathbb{N}^* $
    \FOR{$t \in \{1,...,K\}$}
        \STATE \text{Pull an arm $i_t=t$}
    \ENDFOR
    \FOR{$t \in \{K+1,...,T\}$}
        \FOR{$t \in \{1,...,K\}$}
            \STATE \text{Compute $\hat{R_{t-1}^i}$ and $c_{t-1}^i$} as in (\ref{eq:Rhat}) and (\ref{eq:c})
            \STATE \text{$u_{t-1}^i \leftarrow \hat{R_{t-1}^i} + c_{t-1}^i $}
        \ENDFOR
        \STATE \text{Pull arm $i_t= z = \argmax_{i \in [K]} u_{t-1}^i$ }
        \STATE \text{Observe $x_{h,t-h+1}^{i_h}$ for $h \in \{t-\tau_{max} +1, ..., t\}$}
    \ENDFOR
\end{algorithmic}
\end{algorithm}

% \begin{algorithm}
% \caption{TP-UCB-FR-G}\label{alg:fr}
% \begin{algorithmic}[1]
% \State $\textbf{Input: } a \in [\tau_{max}], \tau_{max} \in \mathbb{N}^* $
% \For{$t \in \{1,...,K\}$}
%     \State \text{Pull an arm $i_t=t$}
% \EndFor
% \For{$t \in \{K+1,...,T\}$}
%     \For{$t \in \{1,...,K\}$}
%         \State \text{Compute $\hat{R_{t-1}^i}$ and $c_{t-1}^i$} as in (\ref{eq:Rhat}) and (\ref{eq:c})
%         \State \text{$u_{t-1}^i \leftarrow \hat{R_{t-1}^i} + c_{t-1}^i $}
%     \EndFor
%     \State \text{Pull arm $i_t= z = \argmax_{i \in [K]} u_{t-1}^i$ }
%      \State \text{Observe $x_{h,t-h+1}^{i_h}$ for $h \in \{t-\tau_{max} +1, ..., t\}$}
% \EndFor
% \end{algorithmic}
% \end{algorithm}
The Generalised Temporally-Partitioned reward UCB with Fictitious Realisations algorithm (see algorithm \ref{alg:fr}) is based on the notion of Fictitious Rewards. With a data distribution function $B()$ as input, the algorithm is able to give a proper judgment of an arm before all the delayed partial rewards are observed. 
As detailed in \Citetext{Romano2022} this is realized by replacing the not yet received partial rewards with fictitious realizations, or in other words, the expected estimated rewards.
At round $t$, the fictitious reward vectors are associated to each arm pulled in the span $H := \{t - \tau_{max} + 1, ..., t - 1 \}$. These fictitious rewards are denoted by $\tilde{\boldsymbol{x}}_h^i=\left[\tilde{x}_{h, 1}^i, \ldots, \tilde{x}_{h, \tau_{\max }}^i\right]$ with $h \in H$, where $\tilde{x}_{h, j}^i:=x_{h, j}^i$, if $h+j \leq t$ (the reward has already been seen), and $\tilde{x}_{h, j}^i=0$, if $h+j>t$ (the reward will be seen in the future). The corresponding fictitious cumulative reward is $\tilde{r}_h^i:=\sum_{j=1}^{\tau_{\max }} \tilde{x}_{h, j}^i$.\\

As input, the algorithm takes a smoothness constant $\alpha \in [\tau_{max}]$, a maximum delay $\tau_{max}$ and a distribution function $)$. The algorithm consists of two phases; the initialization phase, starting in line 2, and the loop phase, starting in line 5. In the initialization phase, each arm is pulled once. After that, the loop phase starts, where at each round $t$, the Upper Confidence Bounds $u_{t-1}^i$ are determined for each arm $i$ by computing the estimated expected reward $\hat{R}_{t-1}^i$ and confidence interval $c_{t-1}^i$, detailed in \ref{theorem:fr}. The algorithm then pulls the arm $i$ with the highest Upper Confidence Bound $u_{t-1}^i$ and observes its rewards.

The estimated expected reward $\hat{R}^i_{t-1}$ is calculated as in \Citetext{Romano2022}.

\begin{equation}
\label{eq:Rhat}
\hat{R}_{t-1}^i:=\frac{1}{N_i(t-1)}\left(\sum_{h=1}^{t-\tau_{\max }} r_h^i \mathbb{I}_{\left\{i_h=i\right\}}+\sum_{h \in H} \tilde{r}_h^i \mathbb{I}_{\left\{i_h=i\right\}}\right)
\end{equation}

where $N_i(t-1):=\sum_{h=1}^{t-1} \mathbb{I}_{\left\{i_h=i\right\}}$ is the number of times arm $i$ has been pulled by the policy up to round $t-1$. \\\\
The confidence term $c^i_{t-1}$ is calculated as:

\begin{equation}
\label{eq:c}
    c_{t-1}^i = 
    \frac{\phi 
    \bar{R}^i}{N_i(t-1)} \sum_{k=1}^{\alpha}
    k  B(k)
    +
    \bar{R}^i\sqrt{
    \frac{2\ln (t-1) \sum_{k=1}^{\alpha}
    \left(B(k)\right)^2
    }{N_i(t-1)}
    }
\end{equation}

\begin{theorem}\label{theorem:fr}
In the TP-MAB setting with $\beta$-spread reward, the pseudo-regret of TP-UCB-FR-G after $T$ rounds is:

\begin{equation}
\begin{aligned}
&\mathcal{R}_T(\mathfrak{U}_{FR-G}) \leq \\
&\sum_{i: \mu_i<\mu^*} \frac{4 \ln T\SIGDOLLAR }{\Delta_i}
\left( 1+ \sqrt{1+ \frac{\Delta_i \phi \SIG}{\bar{R}^i \ln T \sum_{k=1}^{\alpha}
    \left(B(k)\right)^2 }}\right) \\
&+ 2\phi \SIG \sum_{i: \mu_i<\mu^*} \bar{R}^i +\left(1+\frac{\pi^2}{3}\right) \sum_{i: \mu_i<\mu^*} \Delta_i
\end{aligned}
\label{proofFRStep}
\end{equation}
\label{proofFR}
\end{theorem}

Observe that $\SIG=\EX[Y]$, meaning the expected value of our random spread variable $Y$ influences the upper bound of the algorithm. Another interesting factor is $\sum_{k=1}^{\alpha}
    \left(B(k)\right)^2$. This factor can be seen as an approximation of the `Index of Coincidence' between rewards. This determines the probability of two reward points being observed in the same z-group. Its minimal value equals $\frac{1}{\alpha}$ and occurs when the $\alpha$-smoothness property holds (uniform distribution). The value is maximal and equal to 1 if all rewards fall into one z-group. \Citetext{friedman1987index} introduced the `Index of Coincidence' to analyse the distribution of letters in cipher texts in order to find patterns that could help to decipher them.
    \\\\Let us also analyse the resulting upper bound by comparing it to the upper bound found by \Citetext{Romano2022}. $\EX[Y]= \frac{\alpha+1}{2}$ in case of $\alpha$-smoothness. For any other $Y \sim \mathbb{F}$ with $\EX[Y]< \frac{\alpha+1}{2}$ the upper bound on the regret is lower and when  $\EX[Y] > \frac{\alpha+1}{2}$, the upper bound is higher. The `Index of Coincidence' with $\alpha$-smoothness equals its minimal value $\frac{1}{\alpha}$. Hence, using any other non-uniform distribution $\mathbb{F}$ will result in a higher `Index of Coincidence' and higher regret. This means that choosing a $\beta$-spread distribution $\mathbb{F}$ with a low mean and a high coincidence index will result in a better upper bound compared to choosing $\mathbb{F}$ with rewards centered towards the end (high mean) and not spread out (high coincidence index). Let us continue by proving the theorem. 

\subsection{$\beta$-spread regret upper bound}
\begin{proof}
The target of this proof is to derive an upper bound on the expected amount of regret for the algorithm. The approach can be divided into four steps. The first section contains a range of facts that are required for the proof. In the second section, the upper bound on the algorithm is derived by determining an upper bound on the sub-optimal arm pulls $\EX[N_i(t)]$. This process consists of three steps:
\begin{enumerate}
    \item First, we show that the probability that an optimal arm is estimated significantly lower than its mean is bounded by $t^{-4}$ 
    \item Secondly, we show the probability of a sub-optimal arm being estimated significantly higher than its mean is bounded by $t^{-4}$ 
    \item Finally, we evaluate how the algorithm performs in distinguishing the difference between the optimal and sub-optimal means.
\end{enumerate}

These bounds and facts in the preliminary provide enough information to prove that the theorem holds.

\subsubsection{Preliminaries}
The relation between the expected amount of sub-optimal arm pulls and the regret of the algorithm is given by:
\begin{equation}
    \mathcal{R}_T\left(\mathfrak{U}_{\mathrm{FR}}\right)=\sum_{i: \mu_i<\mu^*} \Delta_i \mathbb{E}\left[N_i(T)\right]
\end{equation}
\\
 Let us define the true empirical mean of the cumulative reward of arm $i$ computed over $N_i(t)$ arm pulls:

\begin{equation}
    \hat{R}_t^{i, \text { true }}:=\frac{1}{N_i(t)} \sum_{h=1}^t r_h^i \ind_{\left\{i_h=i\right\}}
\end{equation}

The value above assumes that it is known what the cumulative reward of an arm pull is, even if partial rewards are technically still to come in the future. We bound the difference between the true empirical mean and the observed empirical mean as follows:

\begin{equation}
\hat{R}_t^{i, \text { true }}-\hat{R}_t^i=\frac{1}{N_i(t)} \sum_{h=1}^t \sum_{j=1}^{\tau_{\max }}\left(x_{h, j}^i-\tilde{x}_{h, j}^i\right) \ind_{\left\{i_h=i\right\}}
\end{equation}

\begin{equation}
\leq \frac{1}{N_i(t)} \sum_{h=1}^t \sum_{j=1}^{\tau_{\max }}(x_{h, j}^i-\tilde{x}_{h, j}^i)
=\frac{1}{N_i(t)} \sum_{h=\max \left\{1, t-\tau_{\max }+2\right\}}^t \sum_{j=t-h+2}^{\tau_{\max }} x_{h, j}^i
\label{difftrue}
\end{equation}

% introduce beta spread
\begin{equation}
\leq \frac{1}{N_i(t)} \sum_{k=1}^{\alpha}
k \phi \bar{R}^i  B(k)
\label{ineqbeta}
\end{equation}

% equality of beta spread part
\begin{equation}
= \frac{\phi \bar{R}^i}{N_i(t)} \sum_{k=1}^{\alpha}
k  B(k)
\end{equation}

Step (\ref{difftrue}) states that the difference between the true and observed mean comes from all future rewards that are yet to be observed for a maximum of $\tau_{\max}-1$ arms that have been pulled. The closer to $t$, the more pulled arms with yet-to-be-observed rewards exist. Therefore, this amount can be bounded by looping over all z-groups in (\ref{ineqbeta}) to calculate the maximum reward still to be observed and giving higher weight to late z-groups through index $k$. Furthermore, (\ref{ineqbeta}) holds because of the $\beta$-spread property.
\\

The Chernoff-Hoeffding bound (version of Fact 1 in \Citetext{Auer2002}) is defined as follows: Let $X_1, \ldots, X_n$ be random variables with common range $[0,1]$ such that $\mathbb{E}\left[X_t \mid X_1, \ldots, X_{t-1}\right]=\mu$. Let $S_n=X_1+\cdots+X_n$. Then, for all $a \geq 0$
 
\begin{equation}
   % \mathbb{P}\left\{S_n \geq n \mu+a\right\} \leq e^{-2 a^2 / n} \quad 
   % \text { and } 
    \quad \mathbb{P}\left\{S_n \leq n \mu-a\right\} \leq e^{-2 a^2 / n}
    \label{hoeffdingboundMainPaper}
\end{equation}\\

\subsubsection{Deriving the upper bound}
For more detail, this section of the proof can also be found in the appendix with more sub-steps.

In line with the proof from Auer et al. (2002), the bound of expected sub-optimal arm pulls is bounded using the following summations:

\begin{equation}
    \mathbb{E}\left[N_i(t)\right] \leq \ell+\sum_{t=1}^{\infty} \sum_{s=1}^{t-1} \sum_{s_i=\ell}^{t-1} \mathbb{P}\left\{\left(\hat{R}_{t, s}^*+c_{t, s}^*\right) \leq\left(\hat{R}_{t, s_i}^i+c_{t, s_i}^i\right)\right\}
    \label{ineqAuerMainPaper}
\end{equation}

where ${R}_{t, s}^*$ and $c_{t, s}^*$ are the empirical mean and the confidence term of the optimal arm as computed in the algorithm \texttt{TP-UCB-FR-G}, respectively. The $s$ represents the number of times the arm was pulled up to $t$. ${R}_{t, s}$ and $c_{t, s}$ denote the empirical means and confidence terms for sub-optimal arms.
\\\\
For (\ref{ineqAuerMainPaper}) to hold, one of the following three inequalities have to hold as well \cite{Auer2002}:

\begin{equation}
\hat{R}_{t, s}^* \leq \mu^*-c_{t, s}^*
\label{optimaltoolowMainPaper}
\end{equation}
\begin{equation}
    \hat{R}_{t, s_i}^i \geq \mu_i+c_{t, s_i}^i
\label{badtoohighMainPaper}
\end{equation}
\begin{equation}
    \mu^*<\mu_i+2 c_{t, s_i}^i
\label{badoptimalequalMainPaper}
\end{equation}
\\
Let us pay attention to (\ref{optimaltoolowMainPaper}) first and find the following:

\begin{equation}
    \mathbb{P}\left(\hat{R}_{t, s}^*-\mu^* \leq-c_{t, s}^*\right)=\mathbb{P}\left(\hat{R}_{t, s}^{* \text { true }}-\mu^* \leq-c_{t, s}^*+
    \hat{R}_{t, s}^{* \text { true }}-\hat{R}_{t, s}^*\right)
\end{equation}

% insert prev true vs observed difference into the equation
\begin{equation}
    \leq \mathbb{P}\left(\hat{R}_{t, s}^{* \text { true }}-\mu^* \leq-c_{t, s}^*+
    %insert start
    \frac{\phi 
    \bar{R}^*}{s} \sum_{k=1}^{\alpha}
    k  B(k)
    %end inserting
    \right)
\end{equation}
\begin{equation}    
    =\mathbb{P}\left(s\hat{R}_{t, s}^{* \text { true }} \leq
    %Here the first part of c
    s\mu^*-s\sqrt{
    \frac{2\ln t\sum_{k=1}^{\alpha}
    \left(\bar{R}^*B(k)\right)^2
    }{s}
    }\right)
\end{equation}

\begin{equation}
    \leq \exp \left\{-\frac{
    \left(2\sqrt{
    \frac{2\ln t\sum_{k=1}^{\alpha}
    \left(\bar{R}^*B(k)\right)^2
    }{s}
    }\right)
    ^2 s^2}{
    % chernoff hoefdingss b-a
    \sum_{l=1}^{s}
    \sum_{k=1}^{\alpha}
    \left(\bar{R}^*B(k)\right)^2}\right\} 
    % end of chernoff
    \leq e^{-4 \ln t} \leq t^{-4}
    \label{usehoeffdingMainPaper}
\end{equation}

where we used Hoeffding's inequality (defined in step (\ref{hoeffdingboundMainPaper})) in step (\ref{usehoeffdingMainPaper}) and defined $c_{t, s_i}^i = 
    \frac{\phi 
    \bar{R}^i}{s} \sum_{k=1}^{\alpha}
    k  B(k)
    +
    \bar{R}^i\sqrt{
    \frac{2\ln t\sum_{k=1}^{\alpha}
    \left(B(k)\right)^2
    }{s}
    }
$
\\\\
In a similar way, the bound from (\ref{badtoohighMainPaper}) can be derived:

\begin{equation}
    \mathbb{P}\left(\hat{R}_{t, s_i}^i-\mu_i \geq c_{t, s_i}^i\right) \leq \mathbb{P}\left(\hat{R}_{t, s}^{i, \text { true }}-\mu_i \geq \bar{R}^i \sqrt{\frac{2 \ln t}{\alpha s_i}}\right)
\end{equation}
    
\begin{equation}
    \leq e^{-4 \ln t}=t^{-4}
\end{equation}

where we use the fact that by definition $\hat{R}_{t, s_i}^{i} \leq \hat{R}_{t, s_i}^{i,true}$. All that is left to do is to derive a bound on the probability of (\ref{badoptimalequalMainPaper}). Let us assume that the following holds:
\begin{equation}
    \mu^* \geq \mu^i + 2c^i_{t, s}
\end{equation}

\begin{equation}
    \Delta_i \geq 2\left(
    \frac{\phi 
    \bar{R}^i}{s_i} \sum_{k=1}^{\alpha}
    k  B(k)
    +
    \sqrt{
    \frac{2\ln t\sum_{k=1}^{\alpha}
    \left(\bar{R}^ * B(k)\right)^2
    }{s_i}
    }\right)
\end{equation}

we can rewrite this, see (\ref{upperbounddetails}) for more details, which gives us:

\begin{equation}
\begin{aligned}
& s_i \geq 
    \lvalueleft\\
    & \lvalueright
\end{aligned}
\end{equation}

Therefore, we pick $l$ such that:
\begin{equation}
\begin{aligned}
    & l := \Biggl \lceil \lvalueleft  \\
    &  \lvalueright \Biggr \rceil
\end{aligned}
\end{equation}

which makes sure that for $s_i \geq l$ the inequality in step (\ref{badoptimalequalMainPaper}) is always false. The last couple of steps are similar to the final steps in \Citetext{Romano2022} and theorem 1 of \Citetext{Auer2002}. We get:

\begin{equation}
\begin{aligned}
    & \mathbb{E}\left[N_i(t)\right] \leq \\
    %Insert l with ceil
    & \Biggl \lceil \lvalueleft \\
    & \lvalueright \Biggr \rceil 
    \\ 
&    +
    \sum_{t=1}^{\infty} \sum_{s=1}^{t-1} \sum_{s_i=\ell}^{t-1}\left[\mathbb{P}\left(\hat{R}_{t, s}^*-\mu^* \leq-c_{t, s}^*\right)+\mathbb{P}\left(\hat{R}_{t, s_i}^i-\mu_i \geq c_{t, s_i}^i\right)\right] 
\end{aligned}
\end{equation}

\begin{equation}
\begin{aligned}
    &\leq 
    %INSERT l
    \lvalueleft \\ 
    & \lvalueright \\
    &+1+\sum_{t=1}^{\infty} \sum_{s=1}^{t-1} \sum_{s_i=\ell}^{t-1} 2 t^{-4} \\
\end{aligned}
\end{equation}

\begin{equation}
\begin{aligned}
    &\leq 
    \lvalueleft \\ 
    & \lvalueright \\ 
    & + 1 + \frac{\pi^2}{3}\\
\end{aligned}
\end{equation}
\\
The theorem statement follows by the fact that \\ $\mathcal{R}_T\left(\mathfrak{U}_{\mathrm{FR-G}}\right)=\sum_{i: \mu_i<\mu^*} \Delta_i \mathbb{E}\left[N_i(T)\right]$.

\end{proof}

%\subsection{TP-UCB-EW-G}

%\begin{theorem}
%Here we will try to make EW better....
%
%Sidenote: we will probably not do EW algorithm since we focus on lower bound and FR
%\end{theorem}

\section{Conclusions}
In the TP-MAB setting, an agent will receive subsets of the reward over multiple rounds rather than the entire reward for the arm all at once. The reward structure must be characterized in order to provide algorithms with better asymptotic regret upper bounds than delayed-feedback bandit algorithms. Previous work introduced the $\alpha$-smoothness property to characterize the reward structure. The $\alpha$-smoothness property states that the maximum reward in a group of consecutive rewards cannot exceed a fraction of the maximum reward. However, if the cumulative reward of an arm is not distributed uniformly across rounds, the assumption of $\alpha$-smoothness does not fit well. \\
This paper introduces the $\beta$-spread property, which provides a more generalized way to characterize the reward structure. In the $\beta$-spread property, a probability mass function of any distribution $\mathbb{F}$ is used to determine the maximum reward that can be obtained in a group of consecutive rounds. Depending on the scenario being modeled, the probability mass function can be defined as any distribution, such as a beta-binomial distribution. The maximum reward that can be obtained in group $k$, which consists of $\phi$ rounds, is determined by multiplying the probability $P(X=k)$ by the total cumulative reward of an arm. As a result, the $\beta$-spread property can be used to model scenarios in which the maximum reward in a series of consecutive rewards varies between groups. \\
This paper presents the \texttt{TP-UCB-FR-G} algorithm for the TP-MAB setting, which exploits the $\beta$-spread property. The upper bound for the \texttt{TP-UCB-FR-G} algorithm is $O\left(\sum_{k=1}^{\alpha} \left(B(k)\right)^2 \cdot \ln (T)\right)$, as proved in this paper. It was also demonstrated in this paper that when the $\alpha$-smoothness property is replaced by $\beta$-spread, the upper regret bound can be reduced compared to the algorithm \texttt{TP-UCB-FR} of \Citetext{Romano2022} in case of certain distribution assumptions: distributions with low $\EX[Y]$ and low `Index of Coincidence'. In worse scenarios (high $\EX[Y]$ and index), the nature of the reward distribution forces the algorithm to have a higher upper bound on regret.\\
The $\beta$-spread gives a significant advantage since it generalizes how the total cumulative award is assumed to be distributed across several rounds. Based on prior information about how the cumulative reward is distributed over the rounds, the probability distribution used in the $\beta$-spread property can be determined. Consider a scenario in which the majority of the cumulative reward is observed in the first rounds, and the observed partial reward declines exponentially after that. Then we can incorporate this into the model by specifying a probability mass function that best fits this scenario. \\
Currently, the $\beta$-spread property assumes that the total cumulative reward of an arm is distributed over $\alpha$ groups, each consisting of $\phi$ rounds. In other words, the arbitrary probability distribution in the $\beta$-spread is assumed to have a finite domain of size $\alpha$. Future research could look into whether it is possible to eliminate this assumption. It may be worthwhile to investigate how to replace the $\alpha$-value with a certain distribution rather than a constant value. Another potential future research topic related to the $\alpha$-value is to assign different $\alpha$-values to each arm, rather than having the same $\alpha$ value assigned to each arm. Or one could look at having a completely different distribution per arm. \\
Moreover, future research could look at the maximum time span over which the reward is partitioned, denoted by $\tau_{max}$. Currently, the assumption is that $\tau_{max}$ is fixed and consistent across arms. However, depending on the application, the maximum time span may differ depending on the arm being pulled. Therefore, future research could look into a setting that allows to set a different $\tau_{max}$ for each arm. Furthermore, while the $\tau_{max}$ is currently assumed to be finite, it may be interesting to see what happens if this assumption is removed. This paper made an effort to generalize the distribution of partial rewards the TP-MAB setting, but more research is needed to address the assumptions made. Finally, future work may involve analysing the exact impact and correlation of $\EX[Y]$ and the coincidence index on the regret upper bound of the algorithm \texttt{TP-UCB-FR-G}.

\bibliographystyle{ACM-Reference-Format}
\bibliography{references}


\begin{thebibliography}{9}

%%% the Web address.

\ifx \showCODEN    \undefined \def \showCODEN     #1{\unskip}     \fi
\ifx \showDOI      \undefined \def \showDOI       #1{#1}\fi
\ifx \showISBNx    \undefined \def \showISBNx     #1{\unskip}     \fi
\ifx \showISBNxiii \undefined \def \showISBNxiii  #1{\unskip}     \fi
\ifx \showISSN     \undefined \def \showISSN      #1{\unskip}     \fi
\ifx \showLCCN     \undefined \def \showLCCN      #1{\unskip}     \fi
\ifx \shownote     \undefined \def \shownote      #1{#1}          \fi
\ifx \showarticletitle \undefined \def \showarticletitle #1{#1}   \fi
\ifx \showURL      \undefined \def \showURL       {\relax}        \fi
% invisible to TeX
\providecommand\bibfield[2]{#2}
\providecommand\bibinfo[2]{#2}
\providecommand\natexlab[1]{#1}
\providecommand\showeprint[2][]{arXiv:#2}

\bibitem[Auer et~al\mbox{.}(2002)]%
        {Auer2002}
\bibfield{author}{\bibinfo{person}{Peter Auer}, \bibinfo{person}{Nicolò
  Cesa-Bianchi}, {and} \bibinfo{person}{Paul Fischer}.}
  \bibinfo{year}{2002}\natexlab{}.
\newblock \showarticletitle{Finite-time Analysis of the Multiarmed Bandit
  Problem}.
\newblock \bibinfo{journal}{\emph{Machine Learning 2002 47:2}}
  \bibinfo{volume}{47} (\bibinfo{date}{5} \bibinfo{year}{2002}),
  \bibinfo{pages}{235--256}.
\newblock
Issue 2.
\showISSN{1573-0565}
\urldef\tempurl%
\url{https://doi.org/10.1023/A:1013689704352}
\showDOI{\tempurl}


\bibitem[Bubeck and Cesa{-}Bianchi(2012a)]%
        {DBLP:journals/ftml/BubeckC12}
\bibfield{author}{\bibinfo{person}{S{\'{e}}bastien Bubeck} {and}
  \bibinfo{person}{Nicol{\`{o}} Cesa{-}Bianchi}.}
  \bibinfo{year}{2012}\natexlab{a}.
\newblock \showarticletitle{Regret Analysis of Stochastic and Nonstochastic
  Multi-armed Bandit Problems}.
\newblock \bibinfo{journal}{\emph{Foundations and Trends in Machine Learning}}
  \bibinfo{volume}{5}, \bibinfo{number}{1} (\bibinfo{year}{2012}),
  \bibinfo{pages}{1--122}.
\newblock
\urldef\tempurl%
\url{https://doi.org/10.1561/2200000024}
\showDOI{\tempurl}


\bibitem[Bubeck and Cesa{-}Bianchi(2012b)]%
        {Bubeck-Bianchi2012}
\bibfield{author}{\bibinfo{person}{S{\'{e}}bastien Bubeck} {and}
  \bibinfo{person}{Nicol{\`{o}} Cesa{-}Bianchi}.}
  \bibinfo{year}{2012}\natexlab{b}.
\newblock \showarticletitle{Regret Analysis of Stochastic and Nonstochastic
  Multi-armed Bandit Problems}.
\newblock \bibinfo{journal}{\emph{CoRR}}  \bibinfo{volume}{abs/1204.5721}
  (\bibinfo{year}{2012}).
\newblock
\showeprint[arXiv]{1204.5721}
\urldef\tempurl%
\url{http://arxiv.org/abs/1204.5721}
\showURL{%
\tempurl}


\bibitem[Friedman(1987)]%
        {friedman1987index}
\bibfield{author}{\bibinfo{person}{William~Frederick Friedman}.}
  \bibinfo{year}{1987}\natexlab{}.
\newblock \bibinfo{booktitle}{\emph{The index of coincidence and its
  applications in cryptanalysis}}. Vol.~\bibinfo{volume}{49}.
\newblock \bibinfo{publisher}{Aegean Park Press California}.
\newblock


\bibitem[Joulani et~al\mbox{.}(2013)]%
        {Joulani2013}
\bibfield{author}{\bibinfo{person}{Pooria Joulani}, \bibinfo{person}{András
  György}, {and} \bibinfo{person}{Csaba Szepesvári}.}
  \bibinfo{year}{2013}\natexlab{}.
\newblock \showarticletitle{Online Learning under Delayed Feedback}.
\newblock  (\bibinfo{year}{2013}).
\newblock
\urldef\tempurl%
\url{https://doi.org/10.48550/arXiv.1306.0686}
\showURL{%
\tempurl}


\bibitem[Mandel et~al\mbox{.}(2015)]%
        {mandel2015queue}
\bibfield{author}{\bibinfo{person}{Travis Mandel}, \bibinfo{person}{Yun-En
  Liu}, \bibinfo{person}{Emma Brunskill}, {and} \bibinfo{person}{Zoran
  Popovi{\'c}}.} \bibinfo{year}{2015}\natexlab{}.
\newblock \showarticletitle{The queue method: Handling delay, heuristics, prior
  data, and evaluation in bandits}. In \bibinfo{booktitle}{\emph{Twenty-Ninth
  AAAI Conference on Artificial Intelligence}}.
\newblock


\bibitem[Pike-Burke et~al\mbox{.}(2018)]%
        {pike2018bandits}
\bibfield{author}{\bibinfo{person}{Ciara Pike-Burke}, \bibinfo{person}{Shipra
  Agrawal}, \bibinfo{person}{Csaba Szepesvari}, {and} \bibinfo{person}{Steffen
  Grunewalder}.} \bibinfo{year}{2018}\natexlab{}.
\newblock \showarticletitle{Bandits with delayed, aggregated anonymous
  feedback}. In \bibinfo{booktitle}{\emph{International Conference on Machine
  Learning}}. PMLR, \bibinfo{pages}{4105--4113}.
\newblock


\bibitem[Romano et~al\mbox{.}(2022)]%
        {Romano2022}
\bibfield{author}{\bibinfo{person}{Giulia Romano}, \bibinfo{person}{Andrea
  Agostini}, \bibinfo{person}{Francesco Trovò}, \bibinfo{person}{Nicola
  Gatti}, {and} \bibinfo{person}{Marcello Restelli}.}
  \bibinfo{year}{2022}\natexlab{}.
\newblock \showarticletitle{Multi-Armed Bandit Problem with
  Temporally-Partitioned Rewards: When Partial Feedback Counts}.
\newblock  (\bibinfo{year}{2022}).
\newblock
\urldef\tempurl%
\url{https://trovo.faculty.polimi.it/01papers/romano2022multi.pdf.}
\showURL{%
\tempurl}


\bibitem[Wang et~al\mbox{.}(2021)]%
        {Wang2021}
\bibfield{author}{\bibinfo{person}{Siwei Wang}, \bibinfo{person}{Haoyun Wang},
  {and} \bibinfo{person}{Longbo Huang}.} \bibinfo{year}{2021}\natexlab{}.
\newblock \showarticletitle{Adaptive Algorithms for Multi-armed Bandit with
  Composite and Anonymous Feedback}.
\newblock  (\bibinfo{year}{2021}).
\newblock
\urldef\tempurl%
\url{www.aaai.org}
\showURL{%
\tempurl}


\end{thebibliography}

% \section{References}
% \printbibliography
\clearpage
\onecolumn
\appendix

% \section{Beta-binomial distribution}

% \begin{figure}[h]
%   \includegraphics[width=0.7\linewidth]{betabin.png}
%   \caption{Probability mass function of the beta-binomial distribution: examples of possible reward distributions over 11 time steps}
%   \label{fig:BBD}
% \end{figure}

\section{Proofs}

\subsection*{Theorem~\ref{proofFR}: Deriving the upper bound}
\begin{proof}
In line with the proof from \Citetext{Auer2002}, the bound of expected sub-optimal arm pulls is bounded using the following summations:

\begin{equation}
    \mathbb{E}\left[N_i(t)\right] \leq \ell+\sum_{t=1}^{\infty} \sum_{s=1}^{t-1} \sum_{s_i=\ell}^{t-1} \mathbb{P}\left\{\left(\hat{R}_{t, s}^*+c_{t, s}^*\right) \leq\left(\hat{R}_{t, s_i}^i+c_{t, s_i}^i\right)\right\}
    \label{ineqAuer}
\end{equation}

where ${R}_{t, s}^*$ and $c_{t, s}^*$ are the empirical mean and the confidence term of the optimal arm as computed in the algorithm \texttt{TP-UCB-FR-G}, respectively. The $s$ represents the number of times the arm was pulled up to $t$. ${R}_{t, s}$ and $c_{t, s}$ denote the empirical means and confidence terms for sub-optimal arms.
\\\\
For (\ref{ineqAuer}) to hold, one of the following three inequalities have to hold as well \cite{Auer2002}:

\begin{equation}
\hat{R}_{t, s}^* \leq \mu^*-c_{t, s}^*
\label{optimaltoolow}
\end{equation}
\begin{equation}
    \hat{R}_{t, s_i}^i \geq \mu_i+c_{t, s_i}^i
\label{badtoohigh}
\end{equation}
\begin{equation}
    \mu^*<\mu_i+2 c_{t, s_i}^i
\label{badoptimalequal}
\end{equation}
\\
Let us pay attention to (\ref{optimaltoolow}) first and find the following:

\begin{equation}
    \mathbb{P}\left(\hat{R}_{t, s}^*-\mu^* \leq-c_{t, s}^*\right)=\mathbb{P}\left(\hat{R}_{t, s}^{* \text { true }}-\mu^* \leq-c_{t, s}^*+
    \hat{R}_{t, s}^{* \text { true }}-\hat{R}_{t, s}^*\right)
\end{equation}

% insert prev true vs observed difference into the equation
\begin{equation}
    \leq \mathbb{P}\left(\hat{R}_{t, s}^{* \text { true }}-\mu^* \leq-c_{t, s}^*+
    %insert start
    \frac{\phi 
    \bar{R}^i}{s} \sum_{k=1}^{\alpha}
    k  B(k)
    %end inserting
    \right)
\end{equation}
\begin{equation}    
    =\mathbb{P}\left(s\hat{R}_{t, s}^{* \text { true }} \leq
    %Here the first part of c
    s\mu^*-s\sqrt{
    \frac{2\ln t\sum_{k=1}^{\alpha}
    \left(\bar{R}^*B(k)\right)^2
    }{s}
    }\right)
\end{equation}

\begin{equation}
    \leq \exp \left\{-\frac{
    \left(2\sqrt{
    \frac{2\ln t\sum_{k=1}^{\alpha}
    \left(\bar{R}^*B(k)\right)^2
    }{s}
    }\right)
    ^2 s^2}{
    % chernoff hoefdingss b-a
    \sum_{l=1}^{s}
    \sum_{k=1}^{\alpha}
    \left(\bar{R}^*B(k)\right)^2}\right\} 
    % end of chernoff
    \leq e^{-4 \ln t} \leq t^{-4}
    \label{usehoeffding}
\end{equation}

where we used Hoeffding's inequality (defined in step (\ref{hoeffdingboundMainPaper})) in step (\ref{usehoeffding}) and defined $c_{t, s_i}^i = 
    \frac{\phi 
    \bar{R}^i}{s} \sum_{k=1}^{\alpha}
    k  B(k)
    +
    \bar{R}^i\sqrt{
    \frac{2\ln t\sum_{k=1}^{\alpha}
    \left(B(k)\right)^2
    }{s}
    }
$
\\\\\\
In a similar way, the bound from (\ref{badtoohigh}) can be derived:

\begin{equation}
    \mathbb{P}\left(\hat{R}_{t, s_i}^i-\mu_i \geq c_{t, s_i}^i\right) \leq \mathbb{P}\left(\hat{R}_{t, s}^{i, \text { true }}-\mu_i \geq \bar{R}^i \sqrt{\frac{2 \ln t}{\alpha s_i}}\right)
\end{equation}
    
\begin{equation}
    \leq e^{-4 \ln t}=t^{-4}
\end{equation}\\

where we use that fact that by definition $\hat{R}_{t, s_i}^{i} \leq \hat{R}_{t, s_i}^{i,true}$. All that is left to do is to derive a bound on the probability of (\ref{badoptimalequal}). Let us assume that the following holds:
\begin{equation}
    \mu^* \geq \mu^i + 2c^i_{t, s}
    \label{upperbounddetails}
\end{equation}

\begin{equation}
    \Delta_i \geq 2\left(
    \frac{\phi 
    \bar{R}^i}{s_i} \sum_{k=1}^{\alpha}
    k  B(k)
    +
    \sqrt{
    \frac{2\ln t\sum_{k=1}^{\alpha}
    \left(\bar{R}^* B(k)\right)^2
    }{s_i}
    }\right)
\end{equation}

\begin{equation}
\frac{\Delta_i}{2} - \frac{\phi 
    \bar{R}^i}{s_i} \sum_{k=1}^{\alpha}
    k  B(k)
    \geq 
    \sqrt{
    \frac{2\ln t\sum_{k=1}^{\alpha}
    \left(\bar{R}^* B(k)\right)^2
    }{s_i}
    }
\end{equation}

\begin{equation}
\begin{aligned}
& \frac{\Delta_i^2}{4} + \frac{\phi^2 
    (\bar{R}^i)^2}{s_i^2} \left(\sum_{k=1}^{\alpha}
    k  B(k)\right)^2
    -
    2\left(\frac{ \Delta_i \phi 
    (\bar{R}^i)}{2 s_i} \left(\sum_{k=1}^{\alpha}
    k  B(k)\right)   \right)
    \geq \frac{2\ln t\sum_{k=1}^{\alpha}
    \left(\bar{R}^*B(k)\right)^2
    }{s_i}
\end{aligned}
\end{equation}

\begin{equation}
\begin{aligned}
& s_i^2\frac{\Delta_i^2}{4} + \phi^2 
    (\bar{R}^i)^2 \left(\sum_{k=1}^{\alpha}
    k  B(k)\right)^2
    -
    2 s_i\bigg(\frac{ \Delta_i \phi
    (\bar{R}^i)}{2     } \left(\sum_{k=1}^{\alpha}
    k  B(k)\right)   
    + 
    \ln t\sum_{k=1}^{\alpha}
    \left(\bar{R}^* B(k)\right)^2
    \bigg) \geq 0
\end{aligned}
\end{equation}

\begin{equation}
\begin{aligned}
& s_i \geq \frac{4}{(\Delta_i)^2} 
    \Bigg(\frac{\Delta_i \bar{R}^i \phi \SIG}{2} 
    + \ln t \SIG 
    + \sqrt{\ln^2 t \left( \SIGDOLLAR \right)^2 
    + \bar{R}^i\Delta_i \phi \ln t \SIG \SIGDOLLAR} \Bigg)
\end{aligned}
\end{equation}

\begin{equation}
\begin{aligned}
& s_i \geq 
    \frac{2 \phi \bar{R}^i \SIG}{\Delta_i}
   +
    \frac{4 \ln t \SIGDOLLAR }{\Delta_i^2}
    +
    4 \frac{\sqrt{\left(\ln t \SIGDOLLAR \right) \left( 1 + \frac{\SIG \Delta \phi \bar{R}^i }{\ln t \SIGDOLLAR}\right)}}{\Delta_i^2}
\end{aligned}
\end{equation}

\begin{equation}
\begin{aligned}
& s_i \geq 
    \frac{2 \phi \bar{R}^i \SIG}{\Delta_i}
    +
    \frac{4 \ln t \SIGDOLLAR }{\Delta_i^2}
    \left( 1 + \sqrt{ 1 + \frac{ \Delta_i \phi \bar{R}^i \SIG}
    {\ln t \SIGDOLLAR}}\right)
\end{aligned}
\end{equation}

Therefore we pick $l$ such that:
\begin{equation}
    l := \left\lceil \lvalue \right\rceil
\end{equation}

which makes sure that for $s_i \geq l$ the inequality in step (\ref{badoptimalequal}) is always false. The last couple of steps are similar to the final steps in Romano et al and theorem 1 of Auer et al. We get:

\begin{equation}
\begin{aligned}
    \mathbb{E}\left[N_i(t)\right] \leq &
    %Insert l with ceil
    \left\lceil \lvalue \right\rceil 
    \\ 
&    +
    \sum_{t=1}^{\infty} \sum_{s=1}^{t-1} \sum_{s_i=\ell}^{t-1}\left[\mathbb{P}\left(\hat{R}_{t, s}^*-\mu^* \leq-c_{t, s}^*\right)+\mathbb{P}\left(\hat{R}_{t, s_i}^i-\mu_i \geq c_{t, s_i}^i\right)\right] 
\end{aligned}
\end{equation}

\begin{equation}
\begin{aligned}
    &\leq 
    %INSERT l
    \lvalue +1+\sum_{t=1}^{\infty} \sum_{s=1}^{t-1} \sum_{s_i=\ell}^{t-1} 2 t^{-4} \\
\end{aligned}
\end{equation}

\begin{equation}
\begin{aligned}
    &\leq \lvalue + 1 + \frac{\pi^2}{3}\\
\end{aligned}
\end{equation}
\\\\
The theorem statement follows by the fact that $\mathcal{R}_T\left(\mathfrak{U}_{\mathrm{FR}}\right)=\sum_{i: \mu_i<\mu^*} \Delta_i \mathbb{E}\left[N_i(T)\right]$.

\end{proof}

\end{document}